\pdfoutput=1

\documentclass[11pt]{article}

\usepackage[]{emnlp2022}
\usepackage{subfig}
\usepackage{booktabs}
\usepackage{times}
\usepackage{latexsym}
\usepackage{graphicx}
\usepackage{tabularx}
\usepackage{multirow}

\usepackage[T1]{fontenc}

\usepackage[utf8]{inputenc}

\usepackage{microtype}

\title{EGCR: Explanation Generation for Conversational Recommendation}

\author{Bingbing Wen  \and  Xiaoning Bu  \and  Chirag Shah \\
        University of Washington\\
\texttt{{bingbw,xiaonb,chirags}@uw.edu}}

\begin{document}
\maketitle
\begin{abstract}
Growing attention has been paid in Conversational Recommendation System (CRS), which works as a conversation-based and recommendation task-oriented tool to provide items of interest and explore user preference. However, existing work in CRS fails to explicitly show the reasoning logic to users and the whole CRS still remains a black box. 
Therefore we propose a novel end-to-end framework named Explanation Generation for Conversational Recommendation (EGCR) based on generating explanations for conversational agents to explain why they make the action. EGCR incorporates user reviews to enhance the item representation and increase the informativeness of the whole conversation. To the best of our knowledge, this is the first framework for explainable conversational recommendation on real-world datasets. 
Moreover, we evaluate EGCR on one benchmark conversational recommendation datasets and achieve comparable performance on both recommendation accuracy and conversation quality than other state-of-the art models. 
Finally, extensive experiments demonstrate that generated explanations are not only having high quality and explainability, but also making CRS more trustworthy. We will make our code available to contribute to the CRS community.
\end{abstract}

\begin{figure}[t]
 \includegraphics[width=0.4\textwidth]{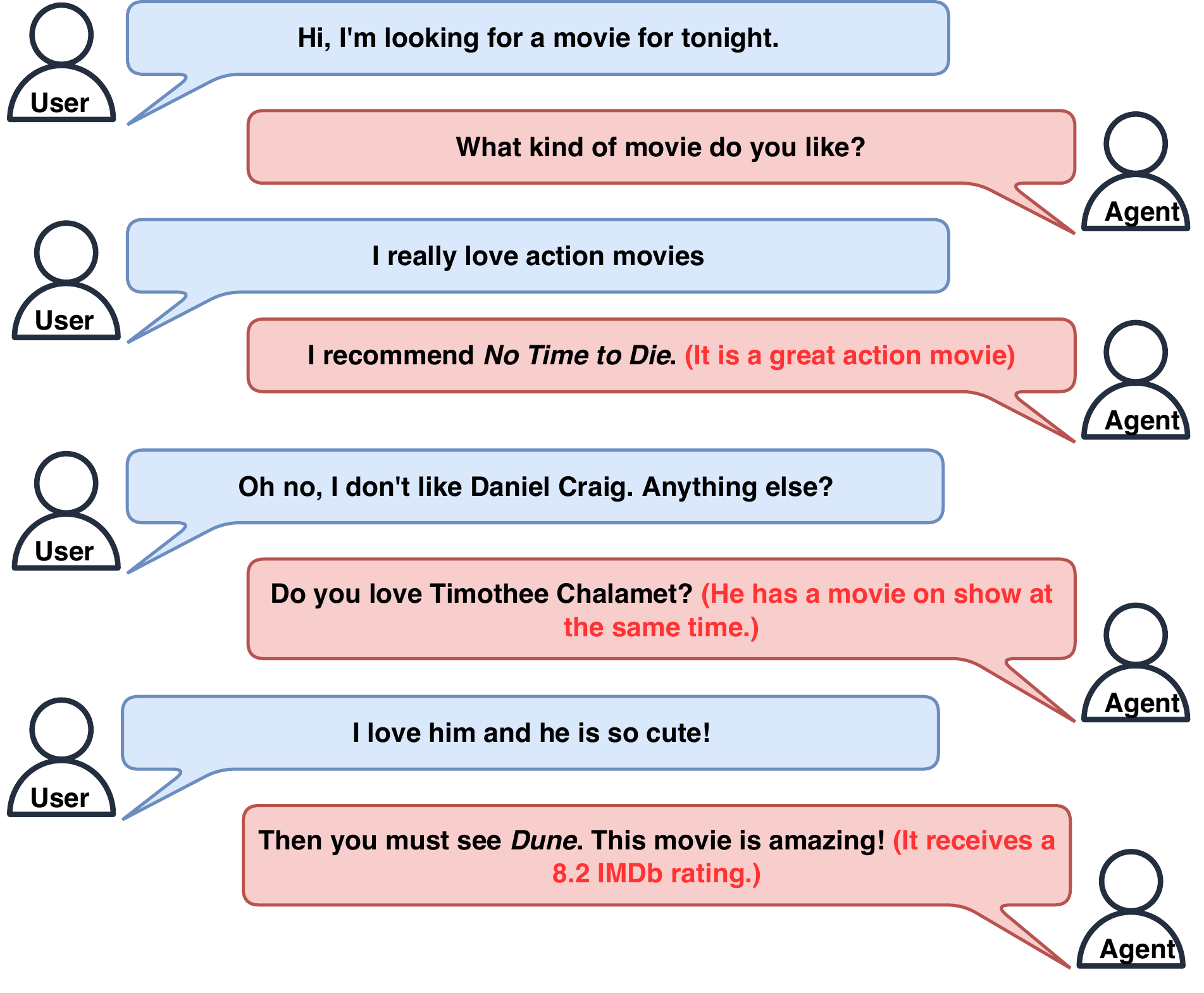}
  \centering
 \caption{An example of explanations for CRS. EGCR will give an explanation for every action. Left: Four turns of conversation with explanations. Red words denote explanations.}
 \label{fig:example}
\vspace{-1cm}
\end{figure}

\section{Introduction}
Conversational agents have been drawing increasing attention as intelligent assistants such as Amazon Alexa, Apple Siri, and Microsoft Cortana have gained immense popularity. Thanks to advances in natural language processing and machine learning, conversational agents are learning to help users finish specific tasks, including searching for information and making recommendations. In particular, growing interest has centered on the Conversational Recommendation System (CRS) that bridges recommendation and conversation.

The design and implementation of Conversational Recommendation System (CRS) is challenging because the interactions behind users, agents, and items are complex and implicit. Existing studies usually focus on using knowledge graphs and context information to achieve better performance on both recommendation and conversation generation~\cite{NEURIPS2018_800de15c,chen2019towards, Zhao2020}. For example, RevCore~\cite{lu-etal-2021-revcore} enriches item information with user reviews to create accurate recommendations and high-quality conversation. CR-Walker~\cite{Ma} proposes tree-structured reasoning over knowledge graphs to guide response generation. These CRS models may vary in detail, but they all remain a black box and fail to give explanations to users why they generate such response and give such recommendation.

In order to better understand the working mechanism and reasoning process behind CRS, we propose a novel framework to generate explanations for CRS. Explanations can make a model more transparent and trustworthy in the explainable recommendation. However, only a few existing works have paid attention to making the conversational recommendation system more explainable. Chen et al.~\cite{ijcai2020-414} incorporated simulated users' feedback through conversations, but it only adopted several pre-defined templates for users' feedback, which may not be comparable to real-world conversation recommendation scenarios.

Figure~\ref{fig:example} shows an example of explanations for CRS. The agent generates explanations for every action. Unlike a single-turn explanation, it integrates context information cross multi-turns in conversation. The agent recommends ``No Time to Die'' because it considers both time ``tonight'' (in Turn 1) and genre ``action'' (in Turn 2). After the user mentions ``Daniel Craig,'' the agent asks a further question about ``Timothee Chalamet'' to explore the user's interests in actors. It then recommended ``Dune'' based on actor and time, which are different reasoning paths on multiple knowledge graphs. The generated explanation demonstrates the reasoning process from ``Daniel Craig'' to  ``Timothee Chalamet.'' Initially, the agent talks about ``Timothee Chalamet'' instead of ``Dune'' to keep the conversation consistent (they talk about actors in the previous turn). 

To address the issue of explainability, we propose a novel framework named Explanation Generation for Conversational Recommendation (EGCR) to generate explanations for every agent action. First, it incorporates user reviews to better represent entities. Then it performs causal reasoning over multiple knowledge graphs (KG), which serve as a background knowledge base. Finally, it generates explanations conditioned on the graph reasoning process and conversation representation.


Our main contributions are as follows:
\begin{itemize}
\item We propose a novel framework named EGCR based on generating explanations for the conversational recommendation, which incorporates user reviews to better explain recommendation. To the best of our knowledge, this is the first framework for explainable conversational recommendation using real-world datasets.
\item Extensive experiments demonstrate that generated explanations not only have high quality and explainability but also make CRS more trustworthy.
\end{itemize}

\begin{figure*}[t]
  \centering
 \includegraphics[width=0.7\textwidth]{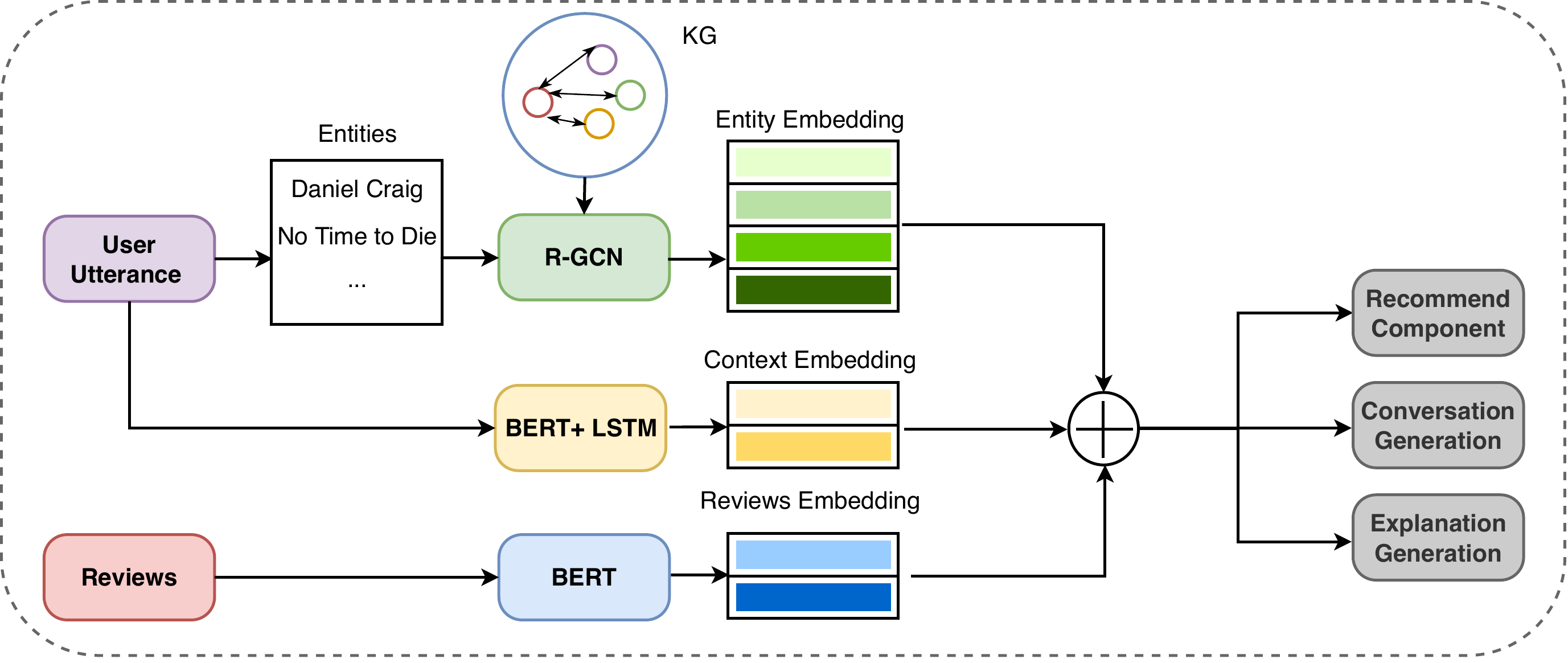}
 \caption{An example of explanations for CRS. EGCR will give an explanation for every action. Left: Four turns of conversation with explanations. Red words denote explanations.}
 \label{fig:framework}
\end{figure*}

\section{Related Work}
A conversational recommendation system (CRS) contains two major components: the recommender module and the conversation module. We categorize the related work into the following three aspects:

\noindent{\textbf{Conversational Recommendation System}} Prior works about conversational recommendation are mainly built upon template-based and review-based question answering pairs~\cite{sun2018conversational,zhang2018personalizing}. These datasets contained structured sentences, such as dialogues between agents and seekers, to clarify personal preferences. 
Kang et al.~\cite{kang-etal-2019-recommendation} collected goal-oriented conversational recommendation. These role-play settings may not reflect real-world scenarios effectively because seekers pretend that they like the given movies. 
Li et al.’s REDIAL dataset~\cite{NEURIPS2018_800de15c} contains chit-chats about movie recommendation. However, recommendations are conditioned based on movies mentioned in the dialog, not on language usage. Moreover, they only mentioned movie names and did not discuss detailed descriptions of movies. Radlinski et al.~\cite{radlinski2019coached} examine movie preference elicitation and Fabian Galetzka~\cite{galetzka2020corpus} examine dialogs about movies. However, these datasets are not conversations about recommendation.

\noindent{\textbf{Explainable Conversational Recommendation}} The landscape of explainable conversational recommendation is scarce.
Most existing works on CRS~\cite{NEURIPS2018_800de15c,lu-etal-2021-revcore, Ma} cannot provide explanations. Because they are not able to use explanations to trigger feedback, these approaches usually adopt questionnaires to collect users' feedback. 
Compared to single-turn explanation generation~\cite{Li2020a,Zhang2018a}, a conversational recommendation system not only involves more items of interest, but also relies on contextual information. Chen et al. ~\cite{ijcai2020-414} is the first work on explainable conversational recommendation. They extend the single-turn explanation generation to a multi-turn explanation generation using template-based feedback simulation. However, simulated user feedback may not reflect real-world scenarios, so we conducted experiments on two real-world conversational recommendation datasets instead of synthetic datasets. 


\section{The EGER Framework}

We construct a knowledge graph $G = (E, R)$, where $E$ represents a set of entities and $\mathbf{R}$ represents the relationships between different entities. Each entity has a set of attributes. There are also relationships between different attributes. Taking movie recommendation as an example, the movie No time to die is linked to attributes like ''Action'' and ''Daniel Craig.'' These two attributes are also linked to ''Genre'' and ''Actor,'' respectively. 



\subsection{Conversation Representation}

\noindent{\textbf{Entity Representation}} We introduce external knowledge like large-scale KG DBpedia~\cite{auer2007dbpedia} and ConceptNet~\cite{speer2017conceptnet} to represent entities. We encode the graph using R-GCN to model neighboring connections more accurately and fuse the different KG by considering different relations.

\noindent{\textbf{Utterance Embedding}} At each dialog turn $t$, we formulate the conversation history $C = {x_1, y_1, . . . , x_{t-1}, y_{t-1}, x_t}$, where $x_t$ and $y_t$ is user/agent utterance respectively. At each conversation turn $t$, we first adopt BERT to encode agent utterance $y_{t-1}$ and  user utterance $x_t$ successively. 
Therefore we could use LSTM over BERT $[y_{t-1}; x_t]$ to encode the dependency among sentences. 


\noindent{\textbf{Review Embedding}} We first adopt BERT to encode reviews and get the embedding of reviews. $R = {r_1, r_2, . . . ,r_i}$. We take the average of all reviews embedding for one movie to enrich the entity representation.
\subsection{Explanation Generation}
We generate explanations conditioned on conversation history by GPT-3~\cite{NEURIPS2020_1457c0d6} prompt tuning. We design a prompt focus on generating explanations for conversational recommendation situations to guide GPT-3~\cite{NEURIPS2020_1457c0d6} to generate proper explanations.

\section{Experiments}

In this section, we introduce the datasets and metrics for evaluation. The baselines and implementation details are also presented.

\subsection{Data}
We construct our datasets on top of a benchmark conversational recommendation dataset. ReDial~\cite{NEURIPS2018_800de15c} is collected by crowd-sourcing workers from Amazon Mechanical Turk (AMT). Two workers take the role of either the recommender or seeker. Each conversation mentions at least 4 different movies, which are labeled clearly. 


\begin{table*}[h]
\centering
 \caption{Qualitative examples of conversations and explanations on Redial dataset. Inside () is explanations generated by our model EGCR.}
 \label{tab:qualitative}
 \scalebox{0.9}{
 \begin{tabular}{p{0.03\linewidth}p{0.4\linewidth}p{0.5\linewidth}}
 \toprule
 \toprule
 Turn & User & Agent\\
 \midrule
1& Hi I am looking for a movie like Super Troopers (2001) & You should watch Police Academy (1984). \textbf{(Both movies are comedies with a similar premise, so I think you would enjoy Police Academy.)}\\ 
\\
2& Is that a great one? I have never seen it. I have seen American Pie. I mean American Pie  (1999). & Yes Police Academy  (1984) is very funny and so is Police Academy 2: Their First Assignment (1985). \textbf{(I would recommend Police Academy (1984) to you because it is a very funny movie. I have never seen it myself, but I have heard good things about it.)}\\
 \bottomrule
\end{tabular}}
\end{table*}

\begin{table}[t]
\centering
 \caption{Automatic evaluation of recommendation and conversation tasks on ReDial}
\label{tab:rec}
\scalebox{0.78}{
\begin{tabular}{|*{7}{c|}}
 \toprule
  \toprule
 & \multicolumn{3}{|c}{Recommendation} & \multicolumn{3}{|c|}{Conversation} \\
 \midrule
 & R@1 & R@10 & R@50  & BLEU & Dist2 & Dist3 \\
 \midrule
 \midrule
 Redial & 2.4 & 14.0 & 32.0 & 21.9 & 14.0 & 32.0  \\
 KBRD & 3.1 & 15.0 & 33.6 & 22.8 & 15.0 & 33.6 \\
 KGSF & 3.9 & 18.3 & 37.8 & 18.6 & 18.3 & 37.8 \\
 CRWalker & 4.0  & 18.7 & 37.6 & 28  & 19.2 & 40.8 \\
 RevCore & 6.1 & 23.6 & 45.4 & - & 42.4 & 55.8 \\
  \midrule
 EGCR & 3.8 & 17.8  & 36.1  & - & 19.1  & 40.4\\
 \bottomrule
\end{tabular}}
\end{table}


\subsection{Baselines}
There is no prior work on generating explanations from these two benchmark datasets. Therefore, we decided to compare our approach with a variety of competitive baselines from previous studies on CRS listed as follows:\\
\textbf{Redial}~\cite{NEURIPS2018_800de15c} construct a hierarchical encoder-decoder architecture with an RNN-based sentiment analysis module. \\
\textbf{KBRD}~\cite{chen2019towards} adopts a KG-enhanced user profile and uses contextual items or entities to generate accurate recommendations. \\
\textbf{KGSF}~\cite{Zhao2020} adopts MIM to fuse the semantic space of two knowledge graphs. It combines the representations of words with items as the user embedding for the recommendation. The generation module is a transformer-based encoder and a fused KG decoder. \\
\textbf{RevCore}~\cite{lu-etal-2021-revcore} incorporates reviews seamlessly to enrich item information and assist in generating both coherent and informative responses. \\
\textbf{CR-Walker}~\cite{Ma} performs tree-structured reasoning over a knowledge graph and generates informative dialog acts to guide language generation.

\subsection{Implementation Details}
To introduce external item information, we crawl 30 reviews for each movie from the IMDb website, one of the most popular movie databases. IMDb provides user reviews, ratings, and scores for corresponding movies. We finally select the 30 reviews with the highest ''helpful'' score for each movie to obtain high-quality reviews. We built our framework upon CR-Walker~\cite{Ma}.

\section{Results and Analysis}

This section presents the results from our experiments, along with our evaluations and interpretations.

\subsection{Evaluation on Explanation Generation Task}

Since there is no existing work and ground truth about generating explanations in conversation recommendation settings, we evaluate the generated explanations in widely-accepted natural language generation metrics. In addition, we show some qualitative examples in Table~\ref{tab:qualitative}.

\subsection{Evaluation on Recommendation and Conversation Task}
For the recommendation task, we use Recall@K(Re@1, R@10, R@50) for evaluation. As the results shown in Table~\ref{tab:rec}. Our model EGCR achieves 3.8\%R@1, 17.8\%R@10 and 36.1\%R@50 on recommendation task. For the Conversation  task, we use Dist-n(Dist-2, Dist-3) and BLEU~\cite{papineni-etal-2002-bleu} to measure the conversation diversity and n-gram matching. As the results shown in Table~\ref{tab:rec}. Our framework achieves 19.1 for Dist-2 and 40.4 for Dist-3, which is also comparable with the SOTA models.

\section{Conclusion}

In this paper, we proposed a novel framework named EGCR based on generating explanations for conversational recommendations, which incorporates reviews to better explain recommendation. To the best of our knowledge, this is the first framework for explainable conversational recommendation using real-world datasets. We evaluated EGCR on one benchmark conversational recommendation datasets and achieved better performance on both recommendation accuracy and conversation quality than other state-of-the-art models. Extensive experiments demonstrate that generated explanations not only have high quality and explainability but could also make CRS more trustworthy.
\bibliography{anthology,custom}
\bibliographystyle{acl_natbib}

\end{document}